\documentclass{article}

\PassOptionsToPackage{numbers, compress}{natbib}

\usepackage[final]{want_neurips_2023}

\usepackage[utf8]{inputenc} 
\usepackage[T1]{fontenc}    
\usepackage{hyperref}       
\usepackage{url}            
\usepackage{booktabs}       
\usepackage{amsfonts}       
\usepackage{nicefrac}       
\usepackage{microtype}      
\usepackage{xcolor}         

\usepackage{comment}
\usepackage{multirow}
\usepackage{booktabs}
\usepackage{diagbox}
\usepackage{graphicx}
	
\usepackage{color, colortbl}

\usepackage{amsmath}
\DeclareMathOperator*{\varhat}{\widehat{var}}

\definecolor{mygray}{gray}{0.9} 

\usepackage[linesnumbered,ruled,vlined]{algorithm2e}
\usepackage{algpseudocode}

\usepackage{subcaption} 

\title{Bandit-Driven Batch Selection\\for Robust Learning under Label Noise}

\author{%
  Michal Lisicki \\
  University of Guelph, Vector Institute \\
  \texttt{mlisicki@uoguelph.ca} \\
  \And
  Mihai Nica \\
  University of Guelph, Vector Institute \\
  \texttt{nicam@uoguelph.ca} \\
  \And
  Graham W.~Taylor \\
  University of Guelph, Vector Institute \\
  \texttt{gwtaylor@uoguelph.ca}
}

\begin{document}

\maketitle

\begin{abstract}
We introduce a novel approach for batch selection in Stochastic Gradient Descent (SGD) training, leveraging combinatorial bandit algorithms. Our methodology focuses on optimizing the learning process in the presence of label noise, a prevalent issue in real-world datasets. Experimental evaluations on the CIFAR-10 dataset reveal that our approach consistently outperforms existing methods across various levels of label corruption. Importantly, we achieve this superior performance without incurring the computational overhead commonly associated with auxiliary neural network models. This work presents a balanced trade-off between computational efficiency and model efficacy, offering a scalable solution for complex machine learning applications.
\end{abstract}

\section{Introduction}

As applications increasingly demand larger and more complex deep learning models, the need for efficient training strategies has become paramount. One way to accelerate training and potentially improve model performance is through the use of Curriculum Learning (CL) and adaptive batch selection. These techniques optimize learning by selectively focusing on data samples that are intrinsically rich and informative at the most appropriate stages of the learning process. Such strategies not only accelerate convergence but also enhance the model's ability to generalize~\cite{loshchilov_online_2016, matiisen_teacher-student_2019, song_carpe_2020}.

While many methods use difficulty metrics to select easy, hard, or uncertain instances for training \cite{wang_survey_2022}, a key area lies in handling noisy or mislabeled datasets \cite{song2022learning}. This domain is particularly important for two reasons: a) the impact of batch selection strategies is easily measured, leading to more insightful conclusions; and b) it addresses the prevalent real-world scenarios where data is often sourced from the web \cite{li2017webvision} or crowdsourced \cite{deng2009imagenet}, and a large portion is considered ``unclean''.

Sample selection strategies using auxiliary Deep Neural Networks
(DNN) effectively mitigate the impact of noisy or mislabeled data. However, these approaches incur substantial computational overhead, limiting their scalability~\cite{han_co-teaching_2018, jiang_mentornet_2018, li_dividemix_2020, yu_how_2019}. While alternative methods like SELFIE~\cite{song_selfie_2019} offer computational efficiency, they are under-explored and rely on steps like re-labeling for optimal performance. Meanwhile, the literature on CL and batch selection offers numerous methods for efficient sample selection across diverse domains~\cite{graves_automated_2017,loshchilov_online_2016}. 

This paper introduces a novel approach that synergizes insights from the CL and batch selection literature to enhance efficient sampling schemes, specifically targeting scenarios with prevalent label noise. Our methodology aims to achieve superior performance without the computational burden often associated with deploying additional DNNs, thereby striking a balance between efficacy and computational efficiency. Unlike traditional CL approaches that focus on individual instances or tasks, our method refines the feedback loop from each training iteration to optimize the \emph{selected batch}. This approach is particularly relevant for tackling the challenges posed by the increasing computational complexity and diversity of machine learning applications across various domains.

\section{Background}

\paragraph{Batch selection and curriculum learning} CL
\cite{bengio_curriculum_2009} and its variants like Self-Paced Learning (SPL)
\cite{kumar_self-paced_2010} and Hard-Example Mining (HEM)
\cite{chang_active_2017,loshchilov_online_2016} provide frameworks for adaptive
instance, batch, or task selection based on difficulty or importance. Despite the efficacy of these
strategies in enhancing stability and convergence, a universal solution remains
elusive, prompting exploration into varied strategies, new importance metrics,
and advanced re-weighting and sampling techniques.

Re-weighting the model's loss by instance, akin to importance-sampling techniques, has been investigated by~\citep{chang_active_2017, loshchilov_online_2016,ren_learning_2018}. 
These studies have indicated that re-weighting can stabilize gradient estimates and reduce bias in the original objective function. However, both \citet{loshchilov_online_2016} and \citet{chang_active_2017} have argued that the impact of this strategy on performance is limited, and that comparable or superior results can be achieved by sampling from a weight-induced distribution. \citet{matiisen_teacher-student_2019} compared the \emph{sample selection} strategies, $\varepsilon$-greedy, Boltzmann, and Thompson sampling, and concluded that the optimal strategy hinges heavily on the sample weight metric. A novel metric not tied to difficulty was introduced by~\citet{chang_active_2017}, emphasizing samples with high prediction uncertainty, and inspired by active learning. 
The authors demonstrated that, by avoiding overly easy or hard instances, their strategy surpassed SPL or HEM on datasets like MNIST and CIFAR with and without label noise. \citet{song_carpe_2020} introduced Recency Bias to boost SGD's convergence by combining principles of~\cite{loshchilov_online_2016} and~\cite{chang_active_2017}. The technique centers on prediction uncertainty, measured by predicted label entropy, as its adaptive sample selection metric. It uses a Boltzmann distribution with energy based on prediction uncertainty and a pressure parameter. Leveraging a sliding window, it emphasizes recent scores, mitigating overfitting and slow convergence. Importantly, Active Bias, Recency Bias, and our approach add minimal computational load to model training.

Automated curriculum learning (ACL) distinguishes itself among CL approaches 
by the degree of control over the learning process. In ACL, the selection 
of tasks is determined dynamically using an algorithm, typically 
an RL or a bandit method. \citet{graves_automated_2017} and~\citet{matiisen_teacher-student_2019} 
have proposed utilizing a non-stationary bandit (Exp3). They demonstrated
that when an agent lacks prior knowledge of its tasks, ACL can significantly 
boost training efficiency relative to uniform sampling. Moreover, a bandit algorithm can discover 
complex orderings and opportunities for efficient knowledge transfer in an unsorted curriculum.
Although prior literature has focused primarily on task-based ACL, the same principles can be utilized in instance and batch selection.

This paper
aims to build upon the mentioned foundational CL techniques by introducing efficient batch
selection methods, particularly in the context of learning with noisy labels.

\paragraph{Efficient learning with noisy labels}


Learning with Noisy Labels (LNL) shares a connection with batch selection but has a different objective. While batch selection picks instances that inherently aid training, LNL focuses on distinguishing between those with clean and noisy labels. Both fields converge when the right batch selection strategy is used to isolate the ``clean'' instances.

LNL is challenging due to DNNs' tendency to memorize complex and possibly incorrect instances, after initially learning simpler patterns~\cite{arpit_closer_2017, zhang_understanding_2017}. This can lead to memorization of inaccurate labels, compromising model generalization. Mislabeled data can also cause confirmation bias, which arises when models overfit to early-selected instances. Multi-network and co-training~\cite{han_co-teaching_2018,yu_how_2019} address these issues but add computational overhead and complexity. Both challenges highlight the need for robust training methods.

The ``small-loss trick'' is a commonly utilized tool for filtering out noisy labels by deeming instances with smaller losses as likely clean. While prevalent in DNN approaches (e.g.,~\cite{han_co-teaching_2018,jiang_mentornet_2018,li_dividemix_2020,shen_learning_2019,yu_how_2019}), 
this method is not optimal when noisy and clean example distributions overlap significantly.
Alternative methods, like measuring prediction uncertainty over time, have been explored as indicators of label corruption~\cite{chang_active_2017,pleiss_identifying_2020,song_selfie_2019}. 

Similarly to batch selection, LNL methods can be broadly categorized into:
loss correction and sample selection~\cite{song2022learning}. Loss correction includes re-weighting or re-labeling. Active Bias~\cite{chang_active_2017} re-weights instances based on prediction variance, bridging LNL and batch selection. A method by~\citet{ren_learning_2018} uses a clean validation set to dynamically assign weights. Re-labeling refines labels from a mix of noisy labels and DNN predictions, as seen in~\cite{reed_training_2015,tanaka_joint_2018,yi_probabilistic_2019}, effectively providing data augmentation. While innovative, these methods pose risks like overfitting to noisy labels and add complexity. Filtering out noisy labels, by contrast, provides a simpler approach but sacrifices information from label refinement.
Sample selection filters out mislabeled data during training, often using an auxiliary DNN. MentorNet~\cite{jiang_mentornet_2018} stands out as a pivotal multi-network approach. It supervises a StudentNet by emphasizing “clean” instances and refining the learning trajectory based on feedback.~\citet{han_co-teaching_2018} proposed a “Co-teaching” paradigm, an alternative, where two DNNs are trained simultaneously and share insights on small-loss instances to diminish errors from noisy labels. “Co-teaching+”~\cite{yu_how_2019} tackles the risk of two networks reaching a consensus with an 'update by disagreement' strategy. The "deep abstaining classifier"~\cite{thulasidasan_combating_2019} is a feature-based multi-network approach, which is particularly effective against structured noise. DivideMix~\cite{li_dividemix_2020} has demonstrated the state-of-the-art LNL performance by employing a semi-supervised (SSL) approach. It dynamically segregates training data into clean and noisy sets using a Gaussian Mixture Model (GMM) and the small-loss trick. To avoid confirmation bias, it utilizes co-teaching. The mislabeled instances are stripped of their labels and refined using SSL~\cite{berthelot2019mixmatch}. Despite their ability to counteract confirmation bias, the multi-network training approaches often come with significant computational overhead.

SELFIE~\cite{song_selfie_2019} is a hybrid approach, combining both loss correction and sample selection. SELFIE seeks to refurbish labels of unclean samples selectively, based on uncertainty, and leverage them along with clean samples, to further reduce false corrections while fully exploiting the current training data. While methods like SELFIE are computationally efficient and provide increase in performance, we argue that their selection strategies can be made better. In this paper we focus on improving the batch selection methodology, and compare performance to the pure selection method, Active Bias~\cite{chang_active_2017}, and two bandit approaches --- the \emph{Exponential-weight algorithm for Exploration and Exploitation} (Exp3)~\cite{auer_using_2002} and \emph{Follow the Perturbed Leader} (FPL)~\cite{lattimore_bandit_2020,neu2016importance}.
Details on these algorithms can be found in Sec.~\ref{sec:methods} and in Appendix,
Sec.~\ref{app:bandit_algorithms}.


\paragraph{Improving sampling efficiency by
exploration}\label{sec:improving-sampling-efficiency-by-exploration}

The challenge of balancing exploration and exploitation is inherent in the
process of batch selection. This balance is crucial for identifying new
instances that can enhance training efficiency and subsequently leverage them
for optimal training outcomes.
In addition, it is essential to account for the high
degree of non-stationarity in neural network training. Specifically, a network
can typically be trained on a particular data instance for only a few iterations
before it risks overfitting. To address this issue, our approach aims to manage
non-stationarity by dynamically adapting weight estimates. This adaptation can
be achieved either through periodic reevaluation, which may be computationally
expensive, or by employing a discounted moving average.

So far we are aware of only one study that has directly compared various
exploration-exploitation strategies in the context of instance selection: its
mixed results suggest such strategies depend highly on the researcher's choice
of a sample weight metric~\citep{matiisen_teacher-student_2019}. Although
$\varepsilon$-greedy and UCB bandit methods have demonstrated effective
performance in instance
selection~\citep{chang_active_2017,matiisen_teacher-student_2019}, the Boltzmann
exploration strategy has recently gained prominence in this
subfield~\citep{cesa-bianchi_boltzmann_2017,loshchilov_online_2016}. In
particular, the adversarial bandit --- Exp3~\citep{auer_using_2002}, which uses
Boltzmann exploration, is commonly utilized as a baseline in non-stationary
environments, and has been shown to be particularly
effective in automated curriculum
learning~\citep{graves_automated_2017,matiisen_teacher-student_2019}.


Our work diverges from prior studies focused on choosing
instances~\cite{chang_active_2017, song_carpe_2020} or tasks~\cite{graves_automated_2017, matiisen_teacher-student_2019}, and targets
batch selection instead.
We utilized the FPL strategy,
which can be thought of as a natural extension of Exp3 into combinatorial (batch),
rather than individual (instance) action selection.

\section{Methods}
\label{sec:methods}

\paragraph{Adversarial multi-armed bandit problem}
A classic baseline approach for non-stationary environments is
the adversarial bandit, in particular, the Exp3 algorithm and its variants. 
In an adversarial $K$-armed bandit problem, at each time step
$t \in \{1, 2, ..., T\}$, the player selects an action
$a_t \in \{1, 2, ..., K\}$ and then an adversary, with full knowledge
of the player's previous actions, assigns a reward
vector $\mathbf{r_t} = (r_{t,1}, r_{t,2}, ..., r_{t,K}) \in [0,1]^K$
across all actions. The player receives a reward $r_{t,a_t}$
corresponding to the selected action $a_t$. There is typically almost
no restrictions on how the adversary can choose
the reward vectors,
as long as the sequence of reward vectors
$\mathbf{r_1}, \mathbf{r_2}, ..., \mathbf{r_T}$ is fixed in advance or
chosen based on the player's past actions. The
player's goal remains to maximize the total collected
reward or equivalently, to minimize regret.

\paragraph{Combinatorial bandits for batch selection}
\label{sec:combinatorial_bandits}
In order to select a full batch of instances at once we need to utilize the
combinatorial bandit paradigm, which considers the joint utility of combinations
of ``basic arms''. Formally, combinatorial bandits can be considered a type of
bandit where a subset of arms is selected in a form of a binary vector
$\mathbf{a}\in\{0,1\}^d$, and the final reward is derived from either a Hadamard or a dot product of that vector with the reward vector $\mathbf{r}$.   
In this work we consider only the subset of a
pre-specified batch size $m$, s.t.~$||\mathbf{a}||_1=m$, and a semi-bandit reward model (see Appendix, Sec.~\ref{app:bandit_algorithms}). A direct application of Exp3 to the semi-bandit problem would entail monitoring
the sequence of estimates for ${K \choose m}$ arms, a task that is
computationally infeasible. The state-of-the-art approach to semi-bandits is
\emph{Follow the Perturbed Leader} (FPL)~\citep{hannan_approximation_1957}, which
mimics Exp3, but estimates probabilities using reward perturbations, rather than
storing them directly.
FPL was originally introduced by~\citet{hannan_approximation_1957} and~\citet{kalai2003cient}, with an efficient version operating on a principle of geometric re-asmpling (GR) proposed by~\citet{neu2016importance}.
In this work, we adapt the FPL algorithm to batch selection.

FPL operates over $n$ rounds, maintaining a vector of weights
$w_{t,i}$ for each action $\mathbf{a}_i$ in the action set $\mathcal{A}$. 
In our case $\mathbf{a}$ is a binary vector, such that setting an $i$-th action $a_i=1$ corresponds to selecting an $i$-th instance $\mathbf{x}_i$. 
Each round the algorithm perturbs the weights with noise $\boldsymbol{\rho}_t$ from distribution $Q$, selecting the action $\mathbf{a}_t$ that maximizes the inner product with the perturbed weight vector. While~\citet{neu2016importance} used the $\text{Exp}(1)$ distribution for $Q$, recent work by~\citet{honda_follow--perturbed-leader_2023} suggests the $\text{Fr\'echet}(2)$ (also known as inverse Weibull) distribution yields optimal regret in adversarial settings. 
As opposed to the algorithms presented in literature \cite{honda_follow--perturbed-leader_2023,neu2016importance} we estimate reward, rather than loss associated with each arm. We have found this adaptation to significantly improve performance in our application, however we acknowledge that while this algorithm remains in line with reward estimation done in Exp3, the original theoretical performance guarantees for combinatorial arm selection may no longer apply.

\begin{algorithm}
\caption{Follow The Perturbed Leader (Reward-guided)}
\KwData{$\mathcal{A}, n, \eta, M, Q$}  

\For{$i = 1$ \KwTo $d$}{
    $w_{0,i} \leftarrow 0$  \tcp{Initialize weight vector}
}

\For{$t = 1$ \KwTo $n$}{
    Sample $\boldsymbol{\rho}_t \sim Q$  \tcp{Sample weight perturbations}
    Compute $\mathbf{a_t} = \arg\max_{\mathbf{a} \in \mathcal{A}} \langle \mathbf{a}, \eta \mathbf{w}_{t-1} + \boldsymbol{\rho}_t \rangle$  \tcp{Choose combinatorial action}

    $\mathbf{r}_{t} \sim \nu_{a_t}$ \tcp{Draw reward vector from arm $\mathbf{a}_t$ of MAB $\nu$}
    
    \ForEach{$i$ with $a_{t,i} = 1$}{ \tcp{Geometric Re-sampling}
        Sample $\sigma_{t,i} \sim \text{Geometric}(p_{t,i})$ \\
        $\hat{r}_{t,i} = \min\{M, \sigma_{t,i}\} a_{t,i}  r_{t,i}$  \tcp{Compute bounded reward estimate}
        $w_{t,i} = w_{t-1,i} + \hat{r}_{t,i}$  \tcp{Update weight of chosen action}
    }
}
\label{alg:fpl}
\end{algorithm}

Following action selection, for each $i$ where
$a_{t,i}=1$, the algorithm proceeds with a geometric re-sampling (GR) step.
Sampling from the geometric distribution estimates $1/p_i$ and in
practice is not done directly, but rather by sampling arms from
$Q + \eta \mathbf{w}_{t-1}$ and counting the number of iterations to
re-occurrance. $M$ is the cap on sampling size, to trade off
computational efficiency with estimation accuracy. The algorithm draws a
sample $\sigma_{t,i}$ from the approximated geometric
distribution, and computes a bounded reward estimate $\hat{r}_{t,i}$
in the same way as Exp3, as an importance-weighted estimate, by
$a_{t,i} \sigma_{t,i} r_{t,i}$.\footnote{$\sigma_{t,i}$ estimates
$1/p_i$}

Finally, the algorithm updates the weight $w_{t,i}$ of the chosen action $a_i$ by
adding the reward estimate $\hat{r}_{t,i}$ to the previous weight $w_{t-1,i}$.
This process continues for $n$ rounds, enabling the algorithm to effectively
explore and exploit the action space by balancing the current estimated rewards
and the exploration noise introduced by perturbations. While FPL may require
more computational resources compared to Exp3, it offers the advantage of
reducing dependency on the combinatorial action space. This makes FPL a
practical choice for real-world sequential decision-making tasks.

\paragraph{Label noise}
According to~\citet{song2022learning}, label noise can be either instance-independent, characterized by constant rates and probabilities, or instance-dependent, where corruption probabilities vary with data features and true labels. This study concentrates on symmetric, instance-independent noise to provide a baseline in a controlled setting.

\paragraph{LNL weight metric}
Choosing the right metric to select informative instances is still an open problem. In the field of LNL, metrics based on prediction loss~\cite{han_co-teaching_2018,jiang_mentornet_2018,li_dividemix_2020,yu_how_2019} and prediction uncertainty~\cite{chang_active_2017,pleiss_identifying_2020,song_selfie_2019} have shown particular promise.
The following metric was proposed
by~\citet{chang_active_2017} as part of the Active Bias method:
$$w_i \propto \varhat(p_{\mathcal{H}^{t-1}_i}(y_i|\mathbf{x}_i)),$$ where, for each
instance $\mathbf{x}_i$, it saves prediction probabilities for their target class over time
in a history buffer $\mathcal{H}_i$, and then computes their variance.

We employed this metric in our study, as it was shown suitable both for LNL and for batch selection in general.
Unlike the metrics that are derived from the change in 
the state of the model, the probability-based metrics reflect the model's current confidence 
in its predictions, rendering them independent of the target solutions, and therefore, 
consistent across instances. This property makes them inherently balanced for problems 
such as LNL. However, it should be noted that while these metrics offer advantages,
they do not directly track the progression of training. Therefore, following \citet{song_carpe_2020}, we limit the size of the history to 10 predictions.


The estimated weights serve two main purposes: either to re-weight the loss as
in \citep{ren_learning_2018} or to parameterize the probability distribution
over data instances. The latter often employs a Boltzmann distribution
(e.g.~\citep{graves_automated_2017,matiisen_teacher-student_2019}): $P_s(i|H, S_e, D) = e^{w_i/\tau} / Z$ where $Z$ is the normalization constant, 
$H$ denotes the history of scores (e.g.~instance losses or prediciton probabilities), $S_e$ is the set of samples used in
the current epoch, and $D = \{(\mathbf{x}_i, y_i) \mid i = 1, 2, \ldots, N\}$ represents
the dataset. Given our interest in the role of exploration in sample efficiency,
we primarily focus on sampling methods underpinned by bandit algorithms such as
Exp3, which also employs a Boltzmann-like distribution.

\section{Results}

\paragraph{Experimental Setup}
We evaluated the performance of various sampling methods including Uniform Sampling, Active Bias, Exp3, and FPL on the CIFAR-10 dataset using a DenseNet model~\cite{huang_densely_2018} with 40 layers. We used the Adam optimizer with momentum 0.9 and an initial learning rate of 0.1 that is decayed by multiplicative factor of 0.1 after 40\,k and 60\,k iterations. The batch size was set to 128 and we ran 200 epochs, consisting of 391 batches each. All methods were repeated 5 times with different seeds, under varying label corruption percentages ranging from 0\% to 50\%. We report the mean and 95\% confidence intervals (CI) of test accuracy achieved by each method.
We will release a PyTorch implementation to reproduce our experiments upon paper acceptance.

\begin{figure}
\centering
\includegraphics[width=1.0\textwidth]{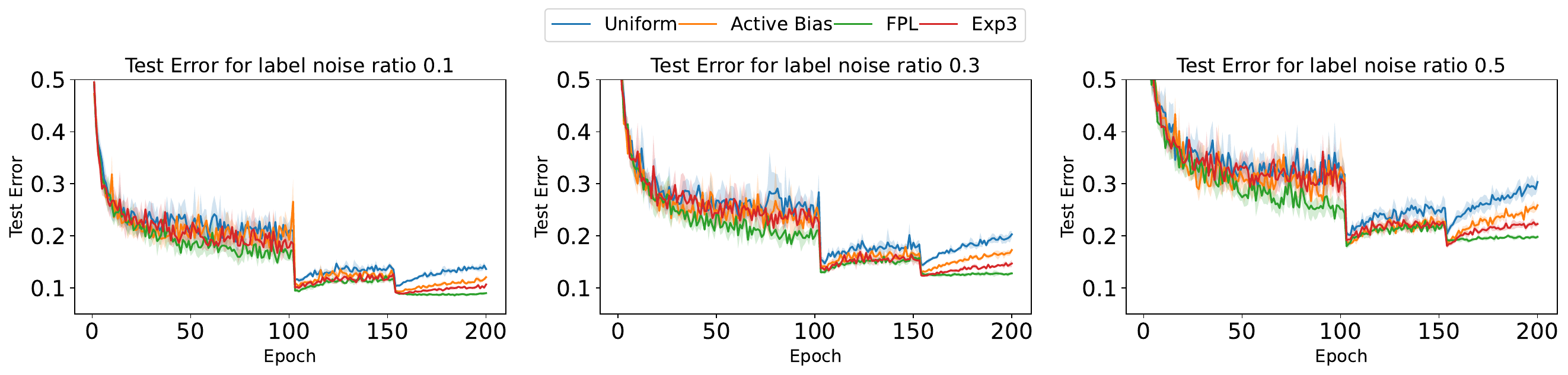}
\caption{Test error over the course of training with confidence intervals (CI) over 5 runs for Uniform, Active Bias (weighted), Exp3 and FPL, for label noise ratio $\in \{0.1, 0.3, 0.5\}$.}
\label{fig:test_curves}
\end{figure}

\paragraph{Results}
Under all label corruption scenarios, FPL exhibited significantly reduced noise and superior performance compared to the other methods, with Exp3 outperforming Active Bias, and Active Bias performing better than Uniform Sampling. In conditions with no label noise, no significant improvement was observed across methods, revealing a potential limitation in sensitivity to ``hard to classify'' instances and an overfocus on mislabeling.

\begin{figure}[!htb]
    \centering
    \begin{minipage}{.3\textwidth}
        \centering
        \tiny
        \setlength{\tabcolsep}{4pt} 
        \begin{tabular}{lllllll}
        \toprule
        \backslashbox{Method}{Noise} & & \multicolumn{1}{c}{10\%} & \multicolumn{1}{c}{20\%} & \multicolumn{1}{c}{30\%} & \multicolumn{1}{c}{40\%} & \multicolumn{1}{c}{50\%} \\
        \midrule
        \rowcolor{mygray}
        \cellcolor{white} \multirow{2}{*}{Uniform} & best & 10.24$\pm$0.25 & 12.05$\pm$0.54 & 14.08$\pm$0.39 & 16.75$\pm$0.30 & 19.40$\pm$0.26 \\
        & last & 13.61$\pm$0.24 & 16.73$\pm$1.30 & 20.28$\pm$0.72 & 23.88$\pm$0.78 & 30.34$\pm$1.70 \\
        \rowcolor{mygray}
        \cellcolor{white} \multirow{2}{*}{Active Bias} & best & 9.17$\pm$0.15 & 10.69$\pm$0.13 & 12.85$\pm$0.09 & 15.09$\pm$0.19 & 18.09$\pm$0.19 \\
        & last & 12.03$\pm$0.57 & 13.74$\pm$0.62 & 17.26$\pm$0.45 & 20.70$\pm$0.31 & 25.86$\pm$0.88 \\
        \rowcolor{mygray}
        \cellcolor{white} \multirow{2}{*}{Exp3} & best & 8.74$\pm$0.13 & 10.28$\pm$0.13 & 12.19$\pm$0.15 & 14.51$\pm$0.23 & 18.02$\pm$0.51 \\
        & last & 10.64$\pm$0.22 & 12.30$\pm$0.49 & 14.67$\pm$0.80 & 17.42$\pm$0.51 & 22.27$\pm$0.50 \\
        \rowcolor{mygray}
        \cellcolor{white} \multirow{2}{*}{FPL} & best & \textbf{8.37$\pm$0.19} & \textbf{10.04$\pm$0.15} & \textbf{12.11$\pm$0.34} & \textbf{14.25$\pm$0.36} & \textbf{17.65$\pm$0.29} \\
        & last & 8.97$\pm$0.15 & 10.46$\pm$0.24 & 12.77$\pm$0.37 & 15.37$\pm$0.46 & 19.79$\pm$0.51 \\
        \bottomrule
        \end{tabular}
    \end{minipage}%
    \hfill 
    \begin{minipage}{.31\textwidth}
        \vspace{1.0em}
        \centering
        \includegraphics[width=\textwidth]{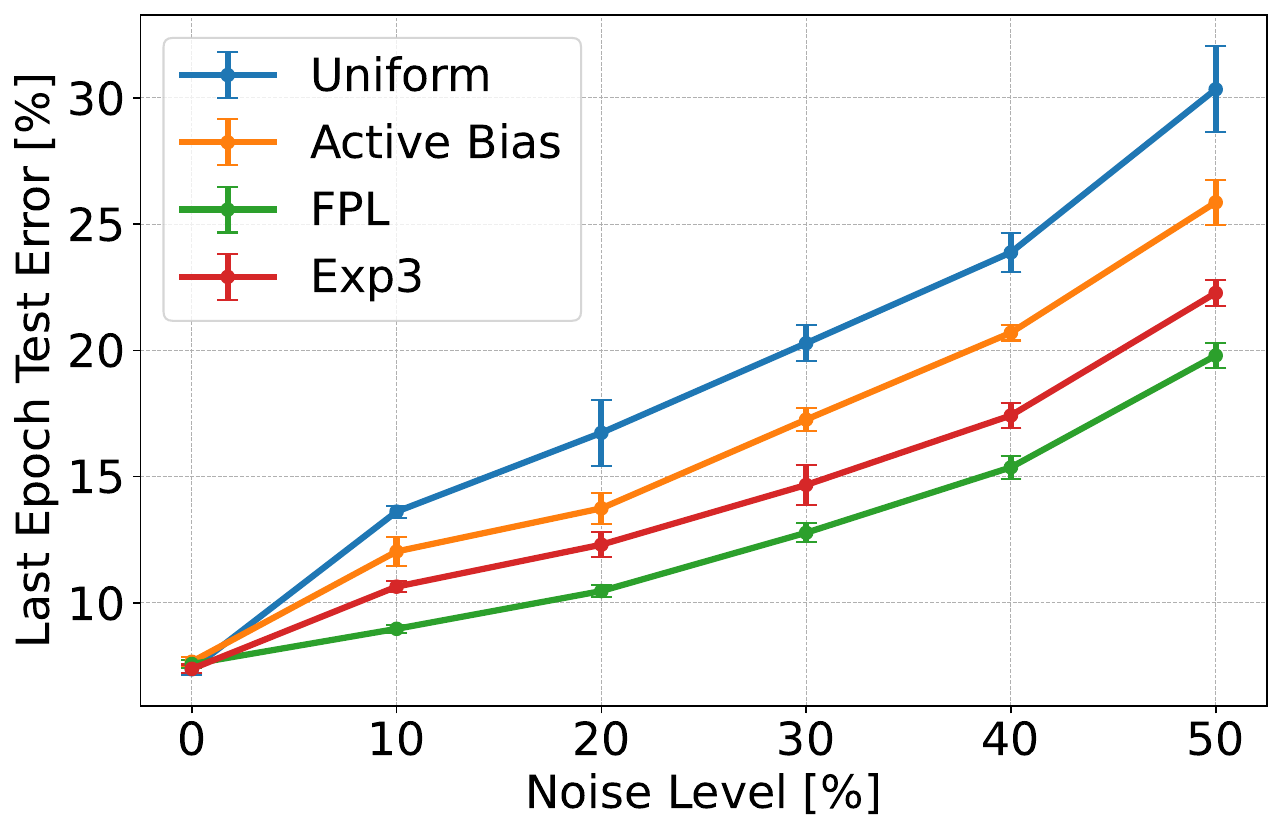}
    \end{minipage}
    \caption{Left: Lowest and final epoch test errors (\%) for each method on CIFAR-10 by noise ratio. Right: Visualizing last epoch performance.}
    \label{fig:scorestable}
\end{figure}

\paragraph{Discussion}
The methods maintained a consistent ranking across noise levels, with the performance gap widening as noise increased (see Fig.~\ref{fig:test_curves} and \ref{fig:scorestable}). FPL consistently yielded smooth and stable convergence, due to its ability to choose informative instances.
When adopting the same weight metric and neural network architecture as Active Bias, our results show that implementing a bandit strategy can lead to significant performance gains. This underscores the importance of not just selecting an optimal weight metric, but also employing a beneficial exploration policy.

\addtocounter{footnote}{-1}
\begin{figure}[ht]
    \centering
    \begin{minipage}[l]{0.38\textwidth}
        \vspace{-0.3em}
        \begin{subfigure}[t]{\textwidth}
            \includegraphics[width=\textwidth]{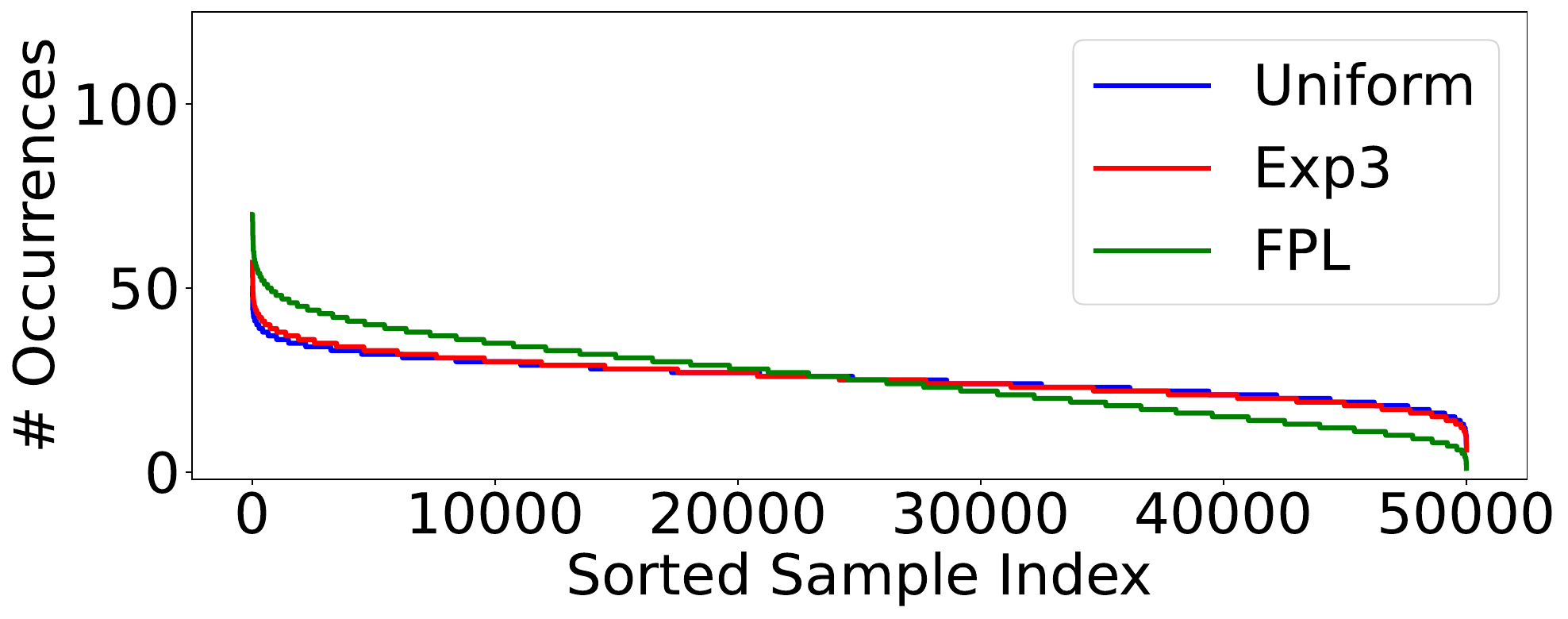}
            \caption{}
            \label{fig:sub1}
        \end{subfigure}
        \begin{subfigure}[b]{\textwidth}
            \includegraphics[width=\textwidth]{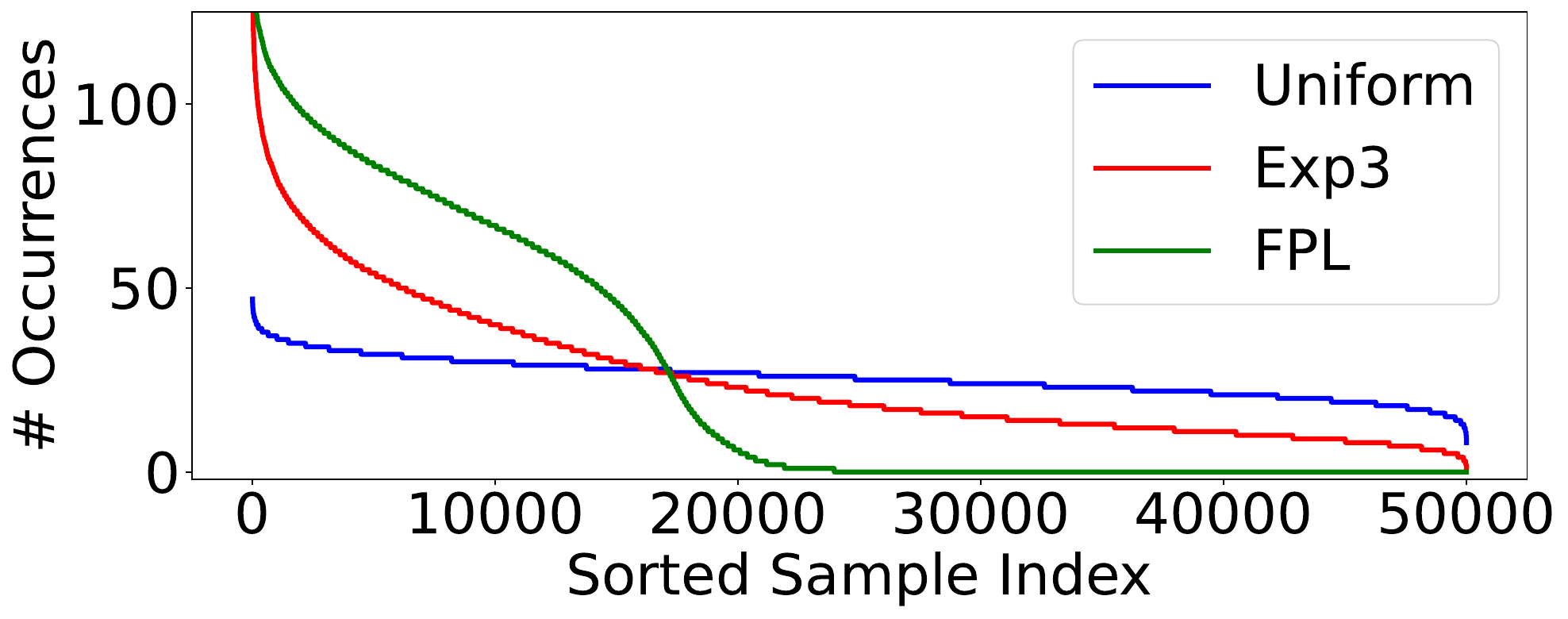}
            \caption{}
            \label{fig:sub2}
        \end{subfigure}
    \end{minipage}%
    \hfill
    \begin{minipage}[r]{0.6\textwidth}
        \hspace{0.5em}
        \begin{subfigure}[r]{\textwidth}
            \includegraphics[width=\textwidth]{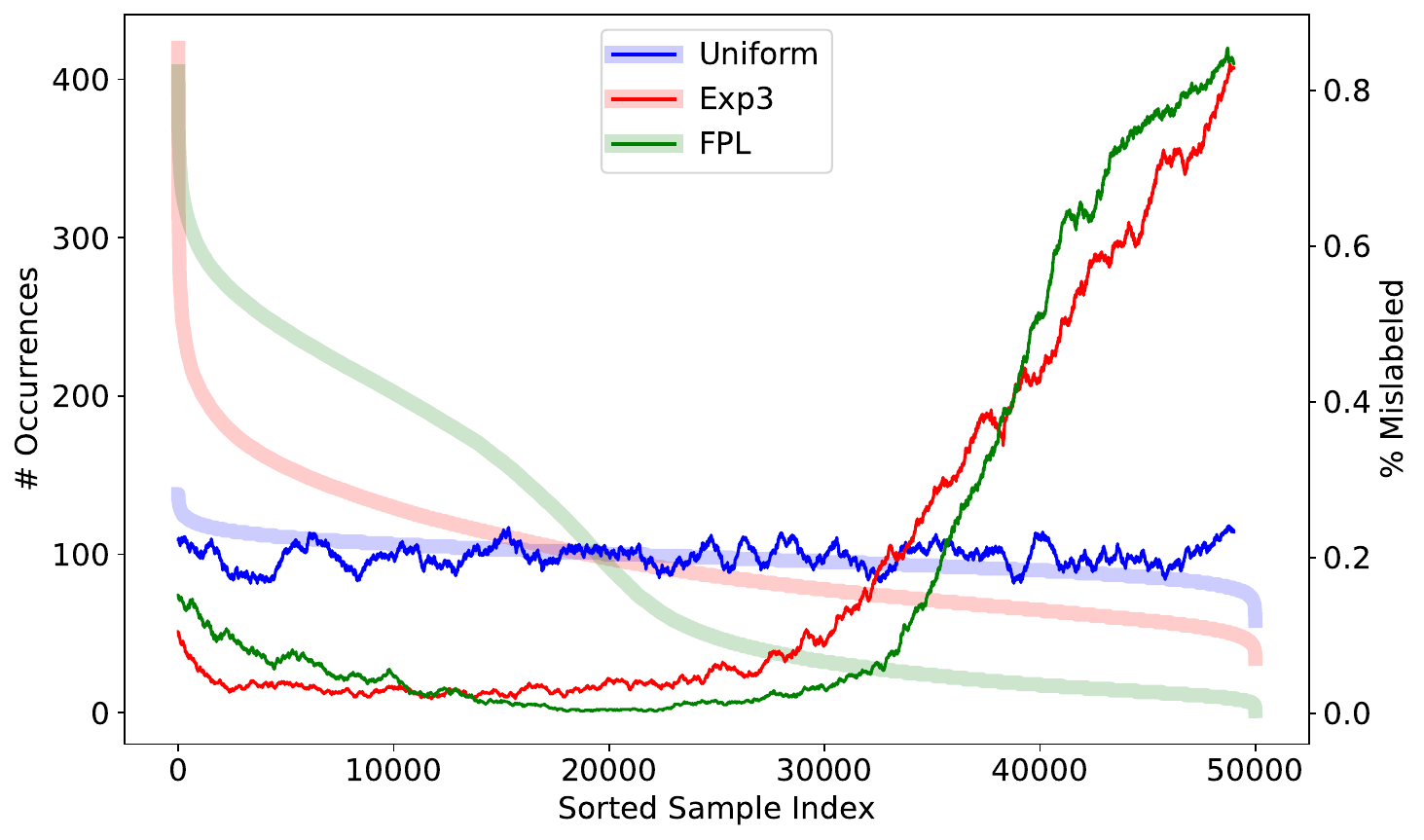}
            \caption{}
            \label{fig:sub3}
        \end{subfigure}
    \end{minipage}
    \caption{Analysis of instance selection for Uniform, Exp3, and FPL\protect\footnotemark\ with 20\% label noise, showing selection occurrence during the initial (a) and final (b) 1000 iterations and total proportion of mislabeled instances (c; front) over total occurrences (c; background). The order of curves at index 0 aligns well with the overall performance of the methods, revealing a concentration of selection in Exp3 and FPL, particularly pronounced in FPL, with Exp3 demonstrating overfitting to a limited set of instances.}
    \label{fig:main}
\end{figure}

Analysis of instance occurrences (Fig.~\ref{fig:main}) reveals insights into the differences in sampling strategies, addressing our initial inquiry into performance gain from utilizing batch- as opposed to instance-based feedback. Initially (Fig.~\ref{fig:main}a), all methods show similar selection frequencies, but distinctions emerge as training concludes (Fig.~\ref{fig:main}b), especially for Exp3 and FPL. Notably, the algorithm's pattern of concentration on specific instances correlates well with its performance. While Exp3 resembles an exponential distribution, FPL produces a threshold at about 20\,k instances, filtering 30\,k of the remaining images. The preference for `clean' instances between the 10\,k and 20\,k sorted index intensifies towards the end of training, indicating the algorithm's inclination to retain instances initially deemed `clean'. These insights emphasize FPL's efficiency as an $m$-set combinatorial bandit method, and highlight its suitability for batch selection. 

\footnotetext{Active bias method is excluded here as we use its loss re-weighting variant, and so its sampling distribution is the same as uniform.}

In Fig.~\ref{fig:main}c, we display curves representing total counts throughout the run, providing a holistic view of each method's sample selection strategy. Over these, with solid lines, we superimpose the percentage of mislabeled instances within a sliding window of 1000 sorted instances. Each point on the overlay represents the mislabeling percentage within that window, revealing a trend: instances sampled less frequently (toward the right) have higher mislabeling percentages. This visualization supports our hypothesis that bandit methods with uncertainty-based metrics, like Exp3 and FPL, enhance performance by focusing on and filtering out mislabeled instances.

While Exp3 effectively identifies mislabeled instances like FPL (Fig.~\ref{fig:main}c), it tends to overfit to a narrow set and over-explore the rest (Fig.~\ref{fig:main}b). This is expected as Exp3 is an instance-based algorithm. However, this overfitting poses risks to its efficacy, as consistently selecting the same subset of impactful instances, combined with a broad array of less pertinent ones, leads to lower performance. Conversely, FPL, by adjusting weights in accordance with other instances, revisits a larger, more balanced subset regularly, forming more informative and consequential batches, ensuring optimal selection in batch training scenarios.

\begin{figure}[ht]
    \centering
    \begin{subfigure}[b]{0.3\textwidth}
        \includegraphics[width=\textwidth]{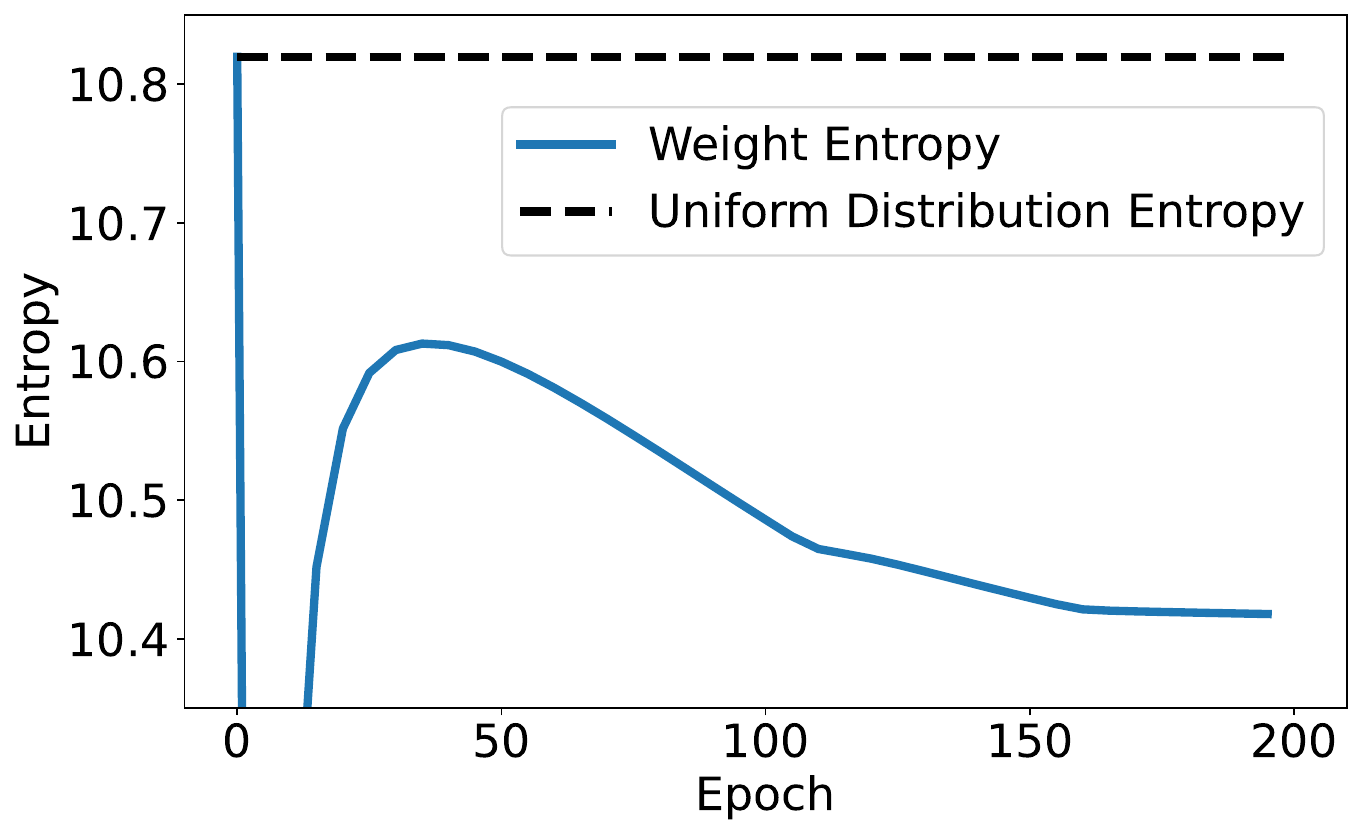}
        \caption{Weight entropy}
        \label{fig:entropy}
    \end{subfigure}
    \hfill
    \begin{subfigure}[b]{0.3\textwidth}
        \includegraphics[width=\textwidth]{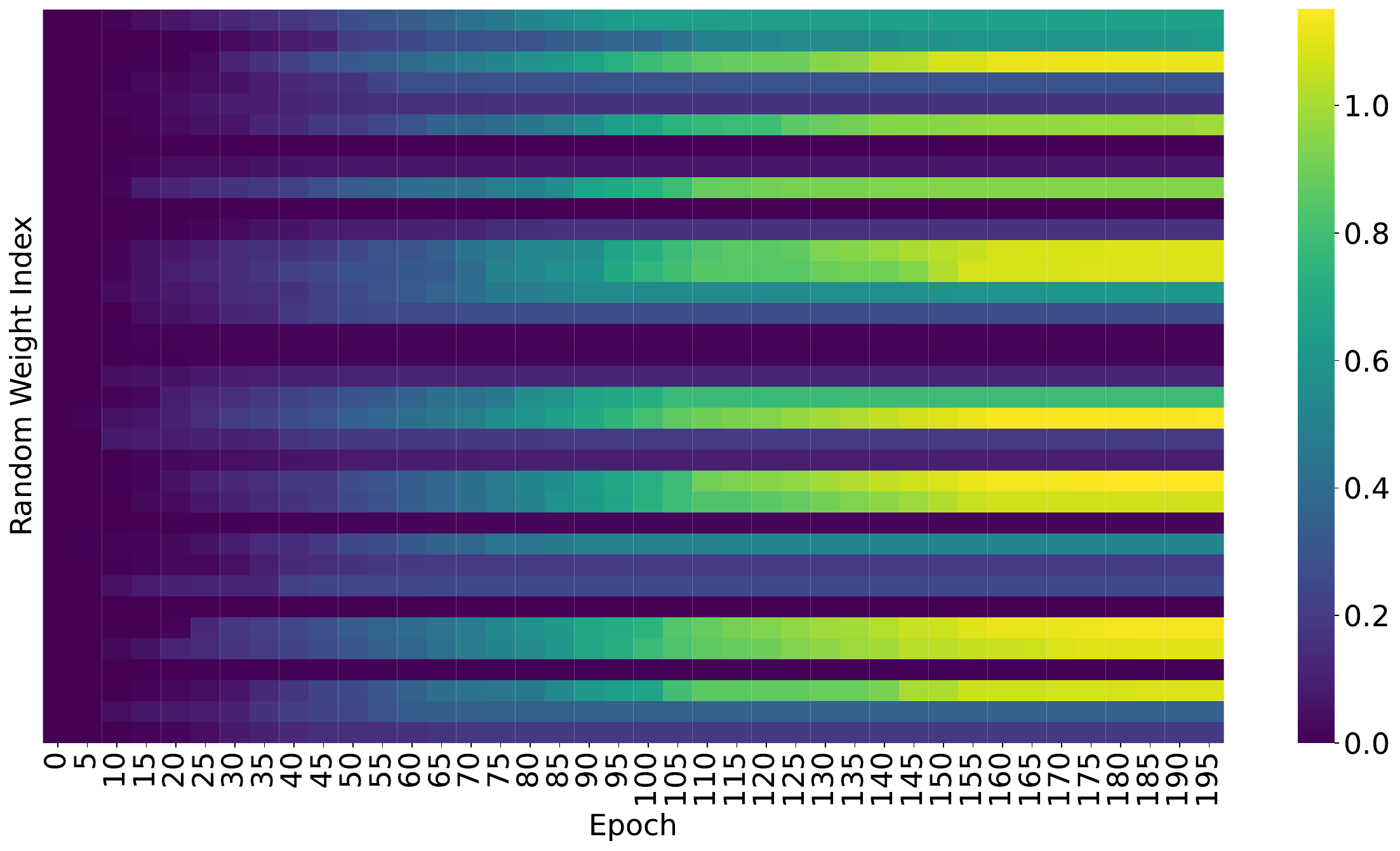}
        \caption{Weights over time}
        \label{fig:weights_over_time}
    \end{subfigure}
    \hfill
    \begin{subfigure}[b]{0.31\textwidth}
        \includegraphics[width=\textwidth]{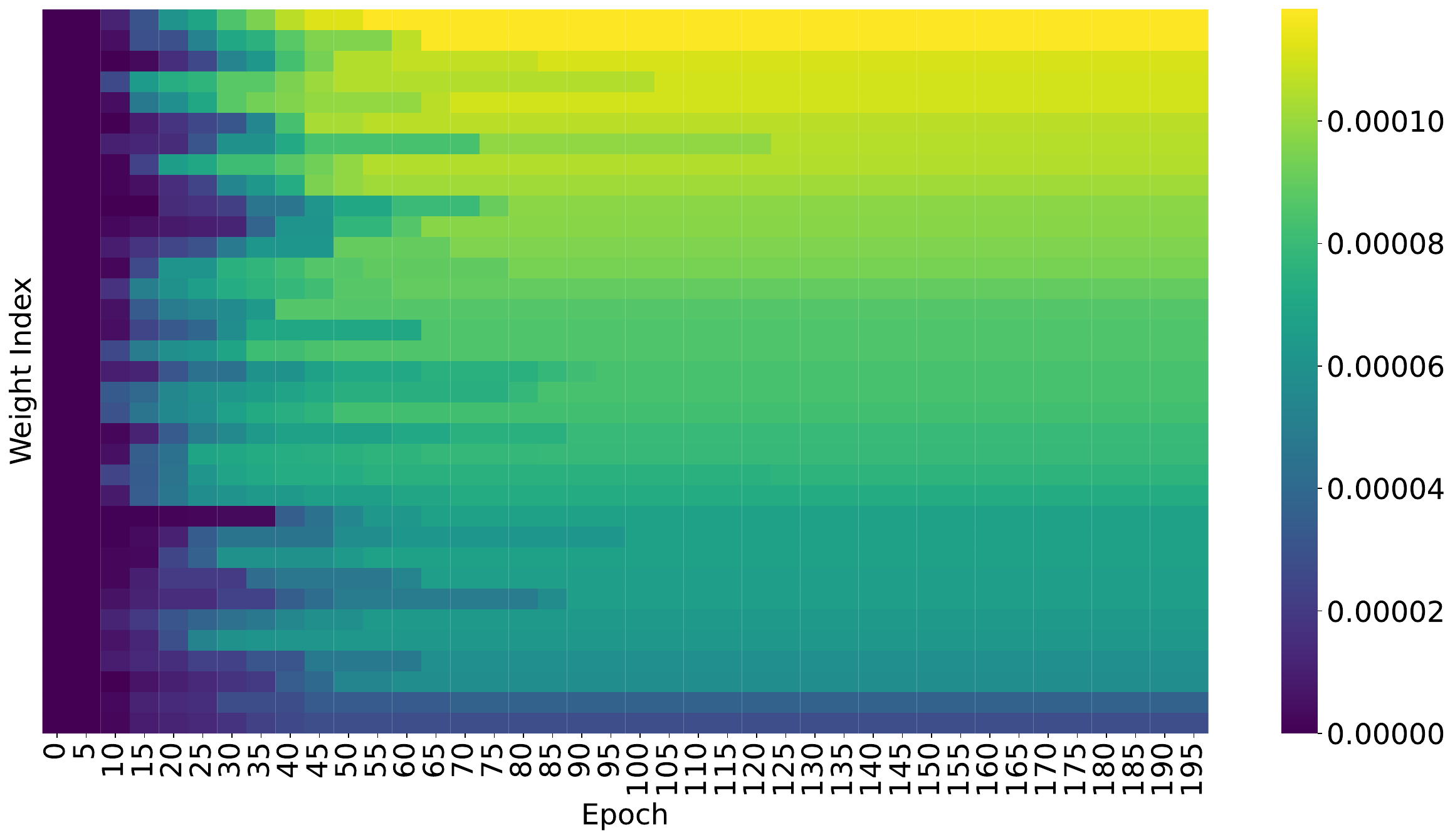}
        \caption{Lowest Weights sorted}
        \label{fig:sorted_wights}
    \end{subfigure}
    \caption{Instance weight visualization. Entropy-aggregated weights are visualized over time in (a). Once all instances are selected by epoch 40, the entropy gradually declines as certain instances gain importance. This pattern indicates effective diversification without overfitting. In (b), a random subset of weights is displayed over time to further validate their individual trajectories. In (c) the lowest weights are shown to dynamics of weights that remain close to 0.}
    \label{fig:weights_entropy}
\end{figure}

As weights play a crucial role in understanding the learning process in depth, we further analyze the weight dynamics of FPL in Fig.~\ref{fig:weights_entropy}. Initially set to 0, the weights adjust smoothly throughout training, maintaining a balance in instance selection without anomalies or overfitting. There is a 40\%-60\% split in instance selection (Fig.~\ref{fig:main}b) that is clearly reflected in the weights, with those corresponding to highly informative instances increasing rapidly, whereas those consistently labeled (either correctly or as mislabeled) remaining closer to zero. It is worth noting here that all instances were selected at least once, with all weights turning strictly positive by the end of the run. Entropy visualization (Fig.~\ref{fig:weights_entropy}a) further emphasizes effective convergence on a well-sized subset of instances, reinforcing selection for better exploitation without excessive exploration.

\begin{figure*}[ht]
    \centering
    \begin{subfigure}[b]{0.3\textwidth}
        \includegraphics[width=\textwidth]{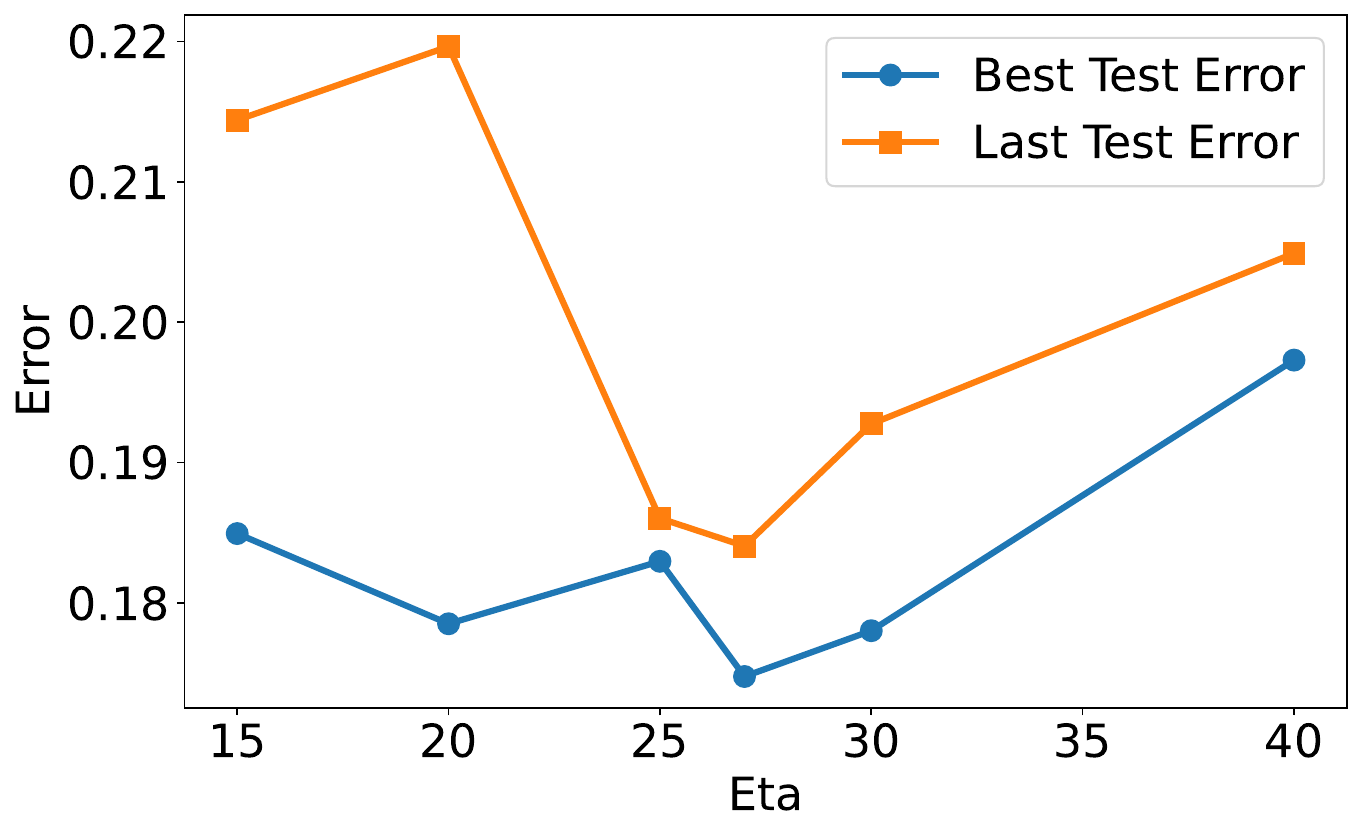}
        \caption{Error w.r.t.~Eta (Beta=25.0, Shape=0.4)}
        \label{fig:sensitivity_eta}
    \end{subfigure}
    \hfill
    \begin{subfigure}[b]{0.3\textwidth}
        \includegraphics[width=\textwidth]{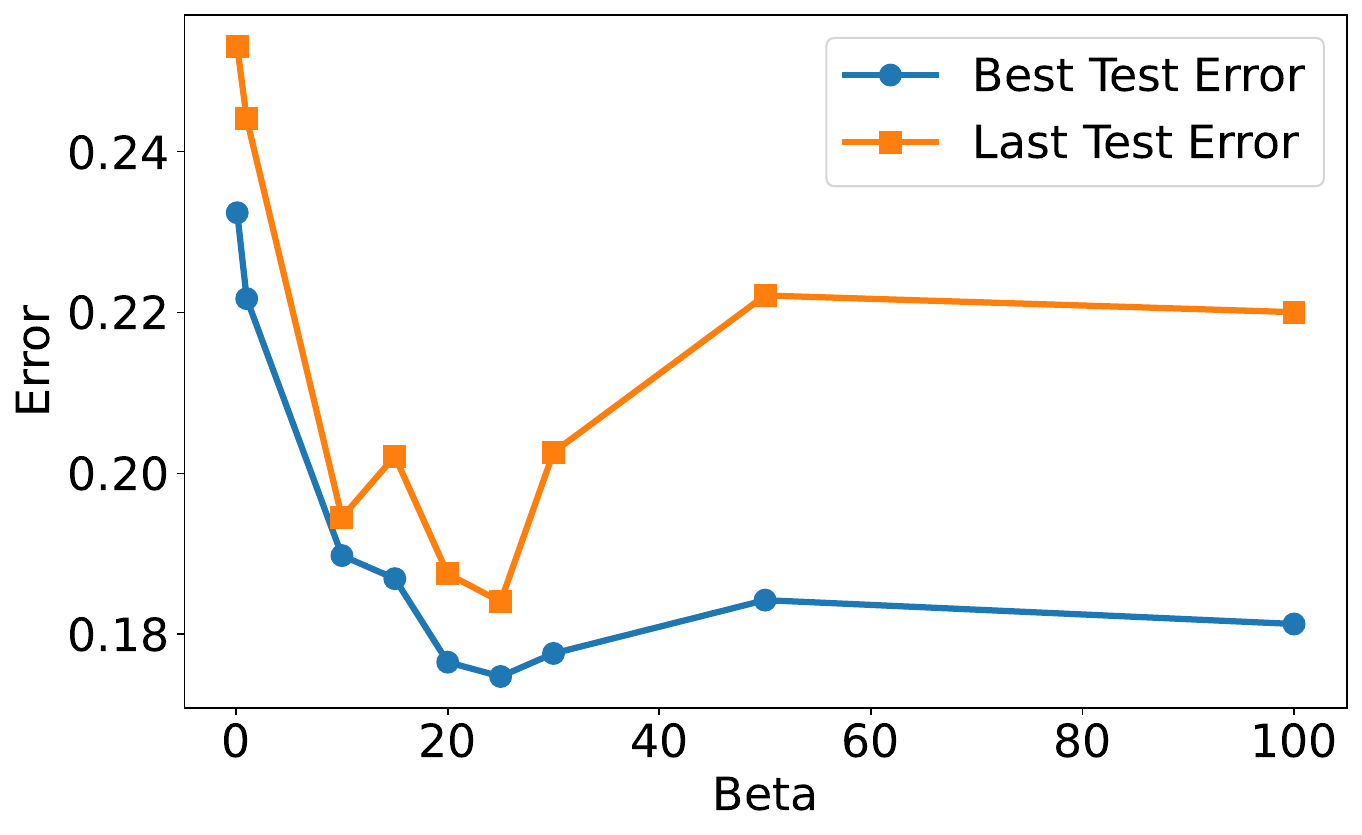}
        \caption{Error w.r.t.~Beta (Eta=27.0, Shape=0.4)}
        \label{fig:sensitivity_beta}
    \end{subfigure}
    \hfill
    \begin{subfigure}[b]{0.3\textwidth}
        \includegraphics[width=\textwidth]{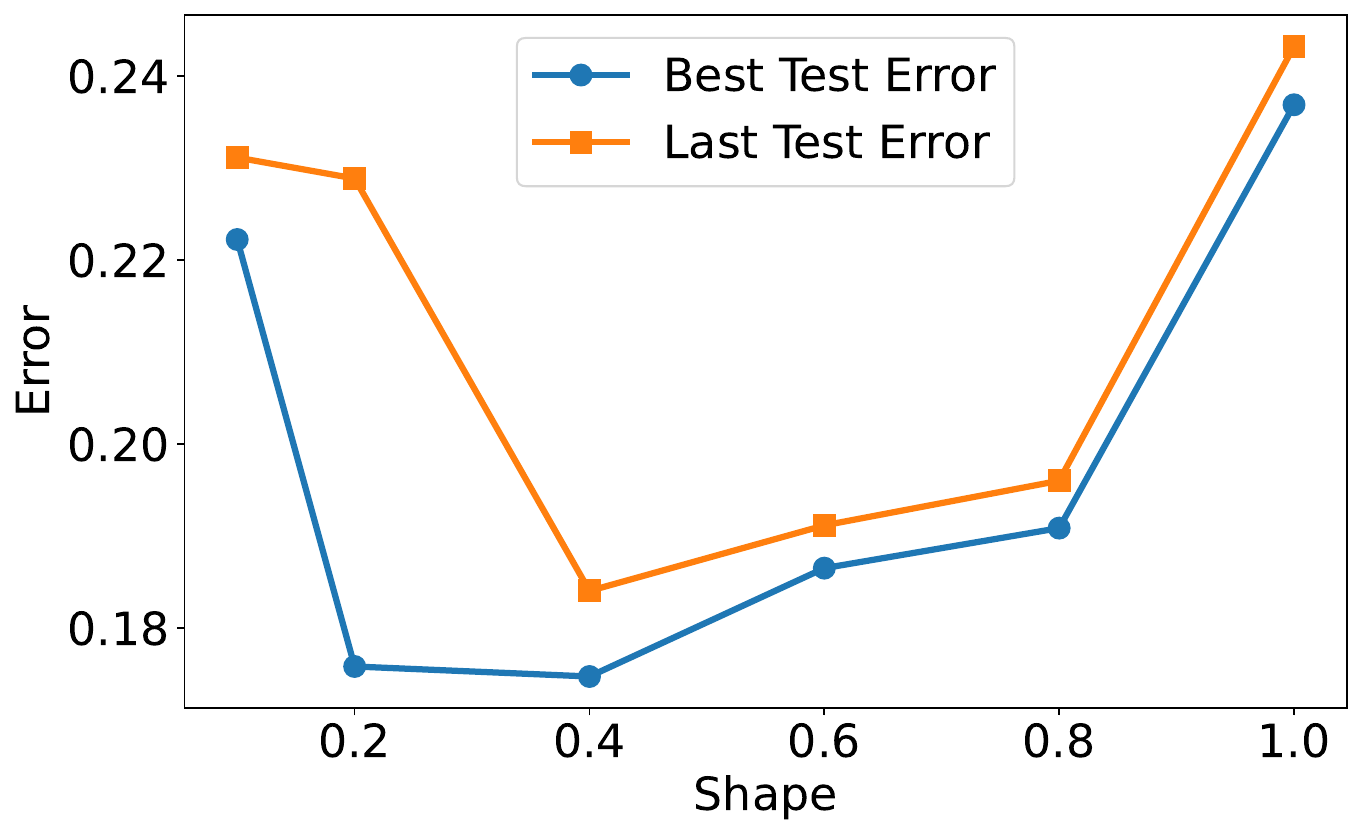}
        \caption{Error w.r.t.~Shape (Eta=27.0, Beta=25.0)}
        \label{fig:sensitivity_shape}
    \end{subfigure}
    \caption{FPL hyperparameter sensitivity analysis for 50\% label noise.}
    \label{fig:sensitivity_analysis}
\end{figure*}

\paragraph{Scalability and hyperparameter sensitivity}
We ran our experiments using $\eta=0.3$ and $\gamma=0.1$ for Exp3 and $\eta \approx 18$, $\beta \approx 20$, and the $\text{Fr\'echet}(0.45)$ distribution for FPL. To show that FPL has small sensitivity to these hyperparameters, we ran a grid search in their vicinity (see Fig.~\ref{fig:sensitivity_analysis}). We found the number of GR samples to be optimal between 500 and 1000. In that range GR introduces an additional computational overhead of 20\%-40\%. This may seem alarming at first, however, we point out that GR is \textit{embarrassingly parallelizable} and instance-based, which makes it scalable in practical applications. 

\paragraph{Limitations and future directions}
Exp3's slower adaptation and potential benefits of its variants like Exp3.P or Exp3.IX warrant further consideration.  
FPL excels in balancing exploration and exploitation but shows limited improvement in noise-free scenarios, suggesting a potential overfocus on mislabeled instances.
While all methods generalize well, tests were limited in scope. Future work will include naturally noisy sets like WebVision~\cite{li2017webvision}, as well as metrics like area under the margin (AUM)~\cite{pleiss_identifying_2020}, to deepen insights and enhance results.


\section{Conclusions}
This investigation into the performance of sampling methods under different noise conditions has revealed key insights into their adaptability, stability, and algorithmic nuances. FPL's effective balance between exploration and exploitation, particularly its focus on uncertain instances, underscores its superior performance. Nonetheless, the absence of marked improvement in noise-free settings and the limited scope of our experiments highlight avenues for future research and refinement.

\bibliographystyle{plainnat}
\bibliography{bibliography}

\begin{thebibliography}{34}
\providecommand{\natexlab}[1]{#1}
\providecommand{\url}[1]{\texttt{#1}}
\expandafter\ifx\csname urlstyle\endcsname\relax
  \providecommand{\doi}[1]{doi: #1}\else
  \providecommand{\doi}{doi: \begingroup \urlstyle{rm}\Url}\fi

\bibitem[Arpit et~al.(2017)Arpit, Jastrzebski, Ballas, Krueger, Bengio, Kanwal,
  Maharaj, Fischer, Courville, Bengio, and Lacoste-Julien]{arpit_closer_2017}
Devansh Arpit, Stanislaw Jastrzebski, Nicolas Ballas, David Krueger, Emmanuel
  Bengio, Maxinder~S. Kanwal, Tegan Maharaj, Asja Fischer, Aaron Courville,
  Yoshua Bengio, and Simon Lacoste-Julien.
\newblock A {Closer} {Look} at {Memorization} in {Deep} {Networks}, July 2017.
\newblock URL \url{http://arxiv.org/abs/1706.05394}.
\newblock arXiv:1706.05394 [cs, stat].

\bibitem[Auer(2002)]{auer_using_2002}
Peter Auer.
\newblock Using confidence bounds for exploitation-exploration trade-offs.
\newblock \emph{Journal of Machine Learning Research}, 3\penalty0
  (Nov):\penalty0 397--422, 2002.

\bibitem[Bengio et~al.(2009)Bengio, Louradour, Collobert, and
  Weston]{bengio_curriculum_2009}
Yoshua Bengio, Jérôme Louradour, Ronan Collobert, and Jason Weston.
\newblock Curriculum learning.
\newblock In \emph{Proceedings of the 26th {Annual} {International}
  {Conference} on {Machine} {Learning}}, {ICML} '09, pages 41--48, New York,
  NY, USA, June 2009. Association for Computing Machinery.
\newblock ISBN 978-1-60558-516-1.
\newblock \doi{10.1145/1553374.1553380}.
\newblock URL \url{https://doi.org/10.1145/1553374.1553380}.

\bibitem[Berthelot et~al.(2019)Berthelot, Carlini, Goodfellow, Papernot,
  Oliver, and Raffel]{berthelot2019mixmatch}
David Berthelot, Nicholas Carlini, Ian Goodfellow, Nicolas Papernot, Avital
  Oliver, and Colin~A Raffel.
\newblock Mixmatch: A holistic approach to semi-supervised learning.
\newblock \emph{Advances in neural information processing systems}, 32, 2019.

\bibitem[Cesa-Bianchi et~al.(2017)Cesa-Bianchi, Gentile, Lugosi, and
  Neu]{cesa-bianchi_boltzmann_2017}
Nicolò Cesa-Bianchi, Claudio Gentile, Gábor Lugosi, and Gergely Neu.
\newblock Boltzmann {Exploration} {Done} {Right}, November 2017.
\newblock URL \url{http://arxiv.org/abs/1705.10257}.

\bibitem[Chang et~al.(2017)Chang, Learned-Miller, and
  McCallum]{chang_active_2017}
Haw-Shiuan Chang, Erik Learned-Miller, and Andrew McCallum.
\newblock Active {Bias}: {Training} {More} {Accurate} {Neural} {Networks} by
  {Emphasizing} {High} {Variance} {Samples}.
\newblock In \emph{Advances in {Neural} {Information} {Processing} {Systems}},
  volume~30. Curran Associates, Inc., 2017.
\newblock URL
  \url{https://proceedings.neurips.cc/paper/2017/hash/2f37d10131f2a483a8dd005b3d14b0d9-Abstract.html}.

\bibitem[Deng et~al.(2009)Deng, Dong, Socher, Li, Li, and
  Fei-Fei]{deng2009imagenet}
Jia Deng, Wei Dong, Richard Socher, Li-Jia Li, Kai Li, and Li~Fei-Fei.
\newblock Imagenet: A large-scale hierarchical image database.
\newblock In \emph{2009 IEEE conference on computer vision and pattern
  recognition}, pages 248--255. Ieee, 2009.

\bibitem[Graves et~al.(2017)Graves, Bellemare, Menick, Munos, and
  Kavukcuoglu]{graves_automated_2017}
Alex Graves, Marc~G. Bellemare, Jacob Menick, Remi Munos, and Koray
  Kavukcuoglu.
\newblock Automated curriculum learning for neural networks.
\newblock In \emph{international conference on machine learning}, pages
  1311--1320. PMLR, 2017.

\bibitem[Han et~al.(2018)Han, Yao, Yu, Niu, Xu, Hu, Tsang, and
  Sugiyama]{han_co-teaching_2018}
Bo~Han, Quanming Yao, Xingrui Yu, Gang Niu, Miao Xu, Weihua Hu, Ivor Tsang, and
  Masashi Sugiyama.
\newblock Co-teaching: {Robust} training of deep neural networks with extremely
  noisy labels.
\newblock In \emph{Advances in {Neural} {Information} {Processing} {Systems}},
  volume~31. Curran Associates, Inc., 2018.
\newblock URL
  \url{https://proceedings.neurips.cc/paper_files/paper/2018/hash/a19744e268754fb0148b017647355b7b-Abstract.html}.

\bibitem[Hannan(1957)]{hannan_approximation_1957}
James Hannan.
\newblock Approximation to {Bayes} risk in repeated play.
\newblock \emph{Contributions to the Theory of Games}, 3:\penalty0 97--139,
  1957.

\bibitem[Honda et~al.(2023)Honda, Ito, and
  Tsuchiya]{honda_follow--perturbed-leader_2023}
Junya Honda, Shinji Ito, and Taira Tsuchiya.
\newblock Follow-the-{Perturbed}-{Leader} {Achieves} {Best}-of-{Both}-{Worlds}
  for {Bandit} {Problems}.
\newblock In \emph{Proceedings of {The} 34th {International} {Conference} on
  {Algorithmic} {Learning} {Theory}}, pages 726--754. PMLR, February 2023.
\newblock URL \url{https://proceedings.mlr.press/v201/honda23a.html}.

\bibitem[Huang et~al.(2018)Huang, Liu, van~der Maaten, and
  Weinberger]{huang_densely_2018}
Gao Huang, Zhuang Liu, Laurens van~der Maaten, and Kilian~Q. Weinberger.
\newblock Densely {Connected} {Convolutional} {Networks}, January 2018.
\newblock URL \url{http://arxiv.org/abs/1608.06993}.
\newblock arXiv:1608.06993 [cs].

\bibitem[Jiang et~al.(2018)Jiang, Zhou, Leung, Li, and
  Fei-Fei]{jiang_mentornet_2018}
Lu~Jiang, Zhengyuan Zhou, Thomas Leung, Li-Jia Li, and Li~Fei-Fei.
\newblock {MentorNet}: {Learning} {Data}-{Driven} {Curriculum} for {Very}
  {Deep} {Neural} {Networks} on {Corrupted} {Labels}, August 2018.
\newblock URL \url{http://arxiv.org/abs/1712.05055}.
\newblock arXiv:1712.05055 [cs].

\bibitem[Kalai and Vempala(2003)]{kalai2003cient}
Adam Kalai and Santosh Vempala.
\newblock E cient algorithms for online decision problems.
\newblock In \emph{Proceedings of the 16th Annual Conference on Computational
  Learning Theory}. Citeseer, 2003.

\bibitem[Kumar et~al.(2010)Kumar, Packer, and Koller]{kumar_self-paced_2010}
M.~Pawan Kumar, Benjamin Packer, and Daphne Koller.
\newblock Self-paced learning for latent variable models.
\newblock In \emph{Proceedings of the 23rd {International} {Conference} on
  {Neural} {Information} {Processing} {Systems} - {Volume} 1}, {NIPS}'10, pages
  1189--1197, Red Hook, NY, USA, December 2010. Curran Associates Inc.

\bibitem[Lattimore and Szepesvári(2020)]{lattimore_bandit_2020}
Tor Lattimore and Csaba Szepesvári.
\newblock \emph{Bandit algorithms}.
\newblock Cambridge University Press, 2020.
\newblock URL \url{https://tor-lattimore.com/downloads/book/}.

\bibitem[Li et~al.(2020)Li, Socher, and Hoi]{li_dividemix_2020}
Junnan Li, Richard Socher, and Steven C.~H. Hoi.
\newblock {DivideMix}: {Learning} with {Noisy} {Labels} as {Semi}-supervised
  {Learning}, February 2020.
\newblock URL \url{http://arxiv.org/abs/2002.07394}.
\newblock arXiv:2002.07394 [cs].

\bibitem[Li et~al.(2017)Li, Wang, Li, Agustsson, and Van~Gool]{li2017webvision}
Wen Li, Limin Wang, Wei Li, Eirikur Agustsson, and Luc Van~Gool.
\newblock Webvision database: Visual learning and understanding from web data.
\newblock \emph{arXiv preprint arXiv:1708.02862}, 2017.

\bibitem[Loshchilov and Hutter(2016)]{loshchilov_online_2016}
Ilya Loshchilov and Frank Hutter.
\newblock Online {Batch} {Selection} for {Faster} {Training} of {Neural}
  {Networks}, April 2016.
\newblock URL \url{http://arxiv.org/abs/1511.06343}.

\bibitem[Matiisen et~al.(2019)Matiisen, Oliver, Cohen, and
  Schulman]{matiisen_teacher-student_2019}
Tambet Matiisen, Avital Oliver, Taco Cohen, and John Schulman.
\newblock Teacher-student curriculum learning.
\newblock \emph{IEEE transactions on neural networks and learning systems},
  2019.

\bibitem[Neu and Bart{{\'o}}k(2016)]{neu2016importance}
Gergely Neu and G{{\'a}}bor Bart{{\'o}}k.
\newblock Importance weighting without importance weights: An efficient
  algorithm for combinatorial semi-bandits.
\newblock \emph{Journal of Machine Learning Research}, 17\penalty0
  (154):\penalty0 1--21, 2016.
\newblock URL \url{http://jmlr.org/papers/v17/15-091.html}.

\bibitem[Pleiss et~al.(2020)Pleiss, Zhang, Elenberg, and
  Weinberger]{pleiss_identifying_2020}
Geoff Pleiss, Tianyi Zhang, Ethan~R. Elenberg, and Kilian~Q. Weinberger.
\newblock Identifying {Mislabeled} {Data} using the {Area} {Under} the {Margin}
  {Ranking}, December 2020.
\newblock URL \url{http://arxiv.org/abs/2001.10528}.
\newblock arXiv:2001.10528 [cs, stat].

\bibitem[Reed et~al.(2015)Reed, Lee, Anguelov, Szegedy, Erhan, and
  Rabinovich]{reed_training_2015}
Scott Reed, Honglak Lee, Dragomir Anguelov, Christian Szegedy, Dumitru Erhan,
  and Andrew Rabinovich.
\newblock Training {Deep} {Neural} {Networks} on {Noisy} {Labels} with
  {Bootstrapping}, April 2015.
\newblock URL \url{http://arxiv.org/abs/1412.6596}.
\newblock arXiv:1412.6596 [cs].

\bibitem[Ren et~al.(2018)Ren, Zeng, Yang, and Urtasun]{ren_learning_2018}
Mengye Ren, Wenyuan Zeng, Bin Yang, and Raquel Urtasun.
\newblock Learning to {Reweight} {Examples} for {Robust} {Deep} {Learning}.
\newblock In \emph{Proceedings of the 35th {International} {Conference} on
  {Machine} {Learning}}, pages 4334--4343. PMLR, July 2018.
\newblock URL \url{https://proceedings.mlr.press/v80/ren18a.html}.

\bibitem[Shen and Sanghavi(2019)]{shen_learning_2019}
Yanyao Shen and Sujay Sanghavi.
\newblock Learning with {Bad} {Training} {Data} via {Iterative} {Trimmed}
  {Loss} {Minimization}, February 2019.
\newblock URL \url{http://arxiv.org/abs/1810.11874}.
\newblock arXiv:1810.11874 [cs, stat].

\bibitem[Song et~al.(2019)Song, Kim, and Lee]{song_selfie_2019}
Hwanjun Song, Minseok Kim, and Jae-Gil Lee.
\newblock {SELFIE}: {Refurbishing} {Unclean} {Samples} for {Robust} {Deep}
  {Learning}.
\newblock In \emph{Proceedings of the 36th {International} {Conference} on
  {Machine} {Learning}}, pages 5907--5915. PMLR, May 2019.
\newblock URL \url{https://proceedings.mlr.press/v97/song19b.html}.
\newblock ISSN: 2640-3498.

\bibitem[Song et~al.(2020)Song, Kim, Kim, and Lee]{song_carpe_2020}
Hwanjun Song, Minseok Kim, Sundong Kim, and Jae-Gil Lee.
\newblock Carpe {Diem}, {Seize} the {Samples} {Uncertain} ``at the {Moment}''
  for {Adaptive} {Batch} {Selection}.
\newblock In \emph{Proceedings of the 29th {ACM} {International} {Conference}
  on {Information} \& {Knowledge} {Management}}, {CIKM} '20, pages 1385--1394,
  New York, NY, USA, October 2020. Association for Computing Machinery.
\newblock ISBN 978-1-4503-6859-9.
\newblock \doi{10.1145/3340531.3411898}.
\newblock URL \url{https://dl.acm.org/doi/10.1145/3340531.3411898}.

\bibitem[Song et~al.(2022)Song, Kim, Park, Shin, and Lee]{song2022learning}
Hwanjun Song, Minseok Kim, Dongmin Park, Yooju Shin, and Jae-Gil Lee.
\newblock Learning from noisy labels with deep neural networks: A survey.
\newblock \emph{IEEE Transactions on Neural Networks and Learning Systems},
  2022.

\bibitem[Tanaka et~al.(2018)Tanaka, Ikami, Yamasaki, and
  Aizawa]{tanaka_joint_2018}
Daiki Tanaka, Daiki Ikami, Toshihiko Yamasaki, and Kiyoharu Aizawa.
\newblock Joint {Optimization} {Framework} for {Learning} with {Noisy}
  {Labels}, March 2018.
\newblock URL \url{http://arxiv.org/abs/1803.11364}.
\newblock arXiv:1803.11364 [cs, stat].

\bibitem[Thulasidasan et~al.(2019)Thulasidasan, Bhattacharya, Bilmes,
  Chennupati, and Mohd-Yusof]{thulasidasan_combating_2019}
Sunil Thulasidasan, Tanmoy Bhattacharya, Jeff Bilmes, Gopinath Chennupati, and
  Jamal Mohd-Yusof.
\newblock Combating {Label} {Noise} in {Deep} {Learning} {Using} {Abstention},
  August 2019.
\newblock URL \url{http://arxiv.org/abs/1905.10964}.
\newblock arXiv:1905.10964 [cs, stat].

\bibitem[Wang et~al.(2022)Wang, Chen, and Zhu]{wang_survey_2022}
Xin Wang, Yudong Chen, and Wenwu Zhu.
\newblock A {Survey} on {Curriculum} {Learning}.
\newblock \emph{IEEE Transactions on Pattern Analysis and Machine
  Intelligence}, 44\penalty0 (9):\penalty0 4555--4576, September 2022.
\newblock ISSN 1939-3539.
\newblock \doi{10.1109/TPAMI.2021.3069908}.

\bibitem[Yi and Wu(2019)]{yi_probabilistic_2019}
Kun Yi and Jianxin Wu.
\newblock Probabilistic {End}-{To}-{End} {Noise} {Correction} for {Learning}
  {With} {Noisy} {Labels}.
\newblock pages 7017--7025, 2019.
\newblock URL
  \url{https://openaccess.thecvf.com/content_CVPR_2019/html/Yi_Probabilistic_End-To-End_Noise_Correction_for_Learning_With_Noisy_Labels_CVPR_2019_paper.html}.

\bibitem[Yu et~al.(2019)Yu, Han, Yao, Niu, Tsang, and Sugiyama]{yu_how_2019}
Xingrui Yu, Bo~Han, Jiangchao Yao, Gang Niu, Ivor~W. Tsang, and Masashi
  Sugiyama.
\newblock How does {Disagreement} {Help} {Generalization} against {Label}
  {Corruption}?, May 2019.
\newblock URL \url{http://arxiv.org/abs/1901.04215}.
\newblock arXiv:1901.04215 [cs, stat].

\bibitem[Zhang et~al.(2017)Zhang, Bengio, Hardt, Recht, and
  Vinyals]{zhang_understanding_2017}
Chiyuan Zhang, Samy Bengio, Moritz Hardt, Benjamin Recht, and Oriol Vinyals.
\newblock Understanding deep learning requires rethinking generalization,
  February 2017.
\newblock URL \url{http://arxiv.org/abs/1611.03530}.
\newblock arXiv:1611.03530 [cs].

\end{thebibliography}


\appendix

\section{Bandit algorithms}
\label{app:bandit_algorithms}

\paragraph{Classic adversarial bandits}


The classic adversarial algorithm, Exp3 (Alg.~\ref{alg:exp3}), employs a
multi-step process for importance adjustments of reward estimates. First, it
adjusts the reward \( r_j \) for arm \( j \) at time \( t \) using the formula
\( \hat{r}_j = \frac{r_{t,j}}{p_{t,j}} \mathbb{I}_{j = i_t} \), where \( p_{t,j} \) is
the probability of choosing arm \( j \). These adjusted rewards are then used to
estimate sample weights \( w_j \). Finally, these weights \( w_j \) are utilized
in the Boltzmann distribution for sampling instances.

\begin{algorithm}
\caption{Exp3 Algorithm}
\KwData{$\gamma\in (0,1]$, $K$} 

\For{$i=1$ \KwTo $K$}{
    $w_{1,i} \leftarrow 1$ \tcp{Initialize weights}
}

\For{each round $t=1,2,...$}{
    $p_{t,j} \leftarrow (1-\gamma)\frac{w_{t,j}}{\sum_{k=1}^K w_{t,k}} +\frac{\gamma}{K}$ \tcp{Compute pmf}
    $i_t \sim p_t$ \tcp{Sample action}
    $r_{t,i_t} \sim \nu_{i_t}$ \tcp{Draw reward from arm $i_t$ of MAB $\nu$}
    $\hat{r}_{j} \leftarrow \frac{r_{t,j}}{p_{t,j}}\mathbb{I}_{\{j=i_{t}\}}$ \tcp{Compute reward estimate}
    $w_{t+1,j} \leftarrow w_{t,j}\exp(\gamma \hat{r}_{t,j})$ \tcp{Update weights}
}
\label{alg:exp3}
\end{algorithm}

The Exp3 algorithm is very efficient computationally and is suitable for task or
individual instance selection, but it doesn't take into account an impact of a
full batch of instances on the performance of the neural network, which may
result in suboptimal performance when competing instances are present in the
same batch.

\paragraph{Analysis and adoption of semi-bandit feedback in combinatorial bandits}

Combinatorial bandits can be categorized as full-information, semi-bandits, or
full-bandit feedback. In the full-bandit feedback scenario we observe just one
reward per batch. While maximizing this reward is our ultimate objective, ignoring the available information about rewards received for
individual arms makes the decision process suboptimal. The full
information setup, where rewards from all the arms are observed is
computationally infeasible, as it requires re-evaluating the network on all the
instances. In our research we adopt the semi-bandit, in which the rewards are
observed only for the basic arms selected in the current round. This setup
aligns well with our problem, where we observe the rewards for instances that
the neural network was trained on in current iteration. As this information is
readily available, no additional passes through the network are required.

Efficient variant of FPL was proposed by~\citet{neu2016importance}, who deployed 
Geometric Re-sampling to estimate probabilities for importance-weighted reward
estimates (re-weighting step), making FPL the first computationally feasible
solution to semi-bandits with strong guarantees. While other methods may offer
comparable or superior theoretical performance in terms of upper regret bounds,
they frequently suffer from computational inefficiency or require additional
optimization steps, rendering them impractical for real-world applications. The
primary appeal of FPL lies in its computational efficiency.

\end{document}